\def\year{2023}\relax
\documentclass[letterpaper]{article} 
\usepackage{aaai23}   
\usepackage{times}  
\usepackage{array}
\usepackage{helvet}  
\usepackage{courier}  
\usepackage[hyphens]{url}  
\usepackage{graphicx} 
\usepackage{natbib}  
\usepackage{caption} 
\DeclareCaptionStyle{ruled}{labelfont=normalfont,labelsep=colon,strut=off} 
\usepackage{algorithm}
\usepackage{algorithmic}
\usepackage[utf8]{inputenc} 
\usepackage[T1]{fontenc}    
\usepackage{hyperref}       
\usepackage{url}            
\usepackage{booktabs}       
\usepackage{amsfonts}       
\usepackage{nicefrac}       
\usepackage{microtype}      
\usepackage{xcolor}         

\usepackage{bm}
\usepackage{subfigure}
\usepackage{wrapfig}
\usepackage{amsmath}
\usepackage{caption}
\usepackage{wrapfig}
\usepackage{subfigure}
\usepackage{amsmath,amssymb}
\usepackage{enumitem}
\usepackage{multirow}
\usepackage{makecell}
\usepackage{newfloat}
\usepackage{listings}

\newcommand{\hankz}[1]{\color{black} #1 \color{black}}

\newcommand{\ignore}[1]{{}}
\floatstyle{ruled}
\newfloat{listing}{tb}{lst}{}
\floatname{listing}{Listing}
\title{A Hierarchical Temporal Planning-Based Approach for Dynamic Hoist Scheduling Problems}
\author{Kebing Jin\textsuperscript{\rm 1}, Yingkai Xiao\textsuperscript{\rm 1}, Hankz Hankui Zhuo\textsuperscript{\rm 1}\thanks{Corresponding author},Renyong Ma\textsuperscript{\rm 2}}

\affiliations{
    \textsuperscript{\rm 1} Sun Yat-Sen University, Guangzhou, China \\

    \textsuperscript{\rm 2} Software Engineering Institute of Guangzhou, Guangzhou, China \\ jinkb@mail2.sysu.edu.cn, xiaoyk8@mail2.sysu.edu.cn, zhuohank@mail.sysu.edu.cn, mry@mail.seig.edu.cn

}

\usepackage{bibentry}

\begin{document}

\newcommand{\tabincell}[2]{\begin{tabular}{@{}#1@{}}#2\end{tabular}}
\def\Domain{$\mathcal{D}$}
\def\SimpleDomain{$\hat{\mathcal{D}}$}
\def\ours{\tt HIT}
\def\pbs{\tt PBS}
\def\shs{\tt SHS}
\def\shsplus{\tt SHS$^+$}
\def\tpp{\tt TPP}

\newcommand{\pre}{\textit{pre}}
\newcommand{\eff}{\textit{eff}}

\newcommand{\product}{\rho}
\newcommand{\Product}{\rho}
\newcommand{\LocationProduct}{l^{\rho}}
\newcommand{\LocationHoist}{l^{\mathcal{H}}}
\newcommand{\Hoist}{\mathcal{H}}
\newcommand{\hoist}{\mathcal{H}}
\newcommand{\Tank}{\mathcal{T}}
\newcommand{\gear}{\mathcal{R}}
\newcommand{\TankRegion}{\mathcal{R}}
\newcommand{\StartT}{ST}
\newcommand{\EndT}{ET}
\newcommand{\StartH}{SH}
\newcommand{\EndH}{EH}
\newcommand{\StartTime}{ \underline{t}}
\newcommand{\EndTime}{ \overline{t}}
\newcommand{\TankTable}{\zeta_t}
\newcommand{\HoistTable}{\zeta}
\newcommand{\tank}{\mathcal{T}}
\newcommand{\Operation}{\mathcal{O}}
\newcommand{\operation}{\mathcal{O}}
\newcommand{\OperationOfTank}{\mathbf{P}}
\newcommand{\duration}{\tau}
\newcommand{\processingtime}{\iota}
\newcommand{\timepoint}{\Upsilon}
\newcommand{\moveduration}{\mathbf{m}}
\newcommand{\updownduration}{\breve{\mathbf{t}}}
\newcommand{\goals}{\mathcal{G}}
\newcommand{\goal}{g}
\newcommand{\Plan}{\Pi}
\newcommand{\plan}{\pi}
\newcommand{\recipe}{\phi}

\maketitle

\begin{abstract}
\hankz{
Hoist scheduling has become a bottleneck in electroplating industry applications with the development of autonomous devices. Although there are a few approaches proposed to target at the challenging problem, they generally cannot scale to large-scale scheduling problems. In this paper, we formulate the hoist scheduling problem as a new temporal planning problem in the form of adapted PDDL, and propose a novel hierarchical temporal planning approach to efficiently solve the scheduling problem. Additionally, we provide a collection of real-life benchmark instances that can be used to evaluate solution methods for the problem. We exhibit that the proposed approach is able to efficiently find solutions of high quality for large-scale real-life benchmark instances, with comparison to state-of-the-art baselines.}
\end{abstract}

\section{Introduction}


New industrial paradigms, such as smart factories and the Internet of Things, are becoming increasingly important with the developments of big data and artificial intelligence, greatly improving the efficiency of manufacturing production processes. These advances allow real-time data to be collected for solving different industrial problems, such as the scheduling of hoists widely used for plating processing, printed circuit boards, etc. Scheduling plays a very important role in automated industrial applications of artificial intelligence \cite{DBLP:journals/ors/Kallrath02,DBLP:journals/ijpr/ParenteFAM20}, aiming to reduce time costs and increase production efficiency.


One mainstream way to handle hoist scheduling problems is to build optimization models according to constraints, and solve them based on mathematical algorithms, e.g., Mixed-Integer Linear Programming (MILP) \cite{DBLP:journals/cor/AmraouiE16,DBLP:journals/candie/FengCC18,DBLP:journals/candie/LaajiliLMN21}. 
However, the uncertainties and large scales let constructing and solving mathematical models  time-consuming and tedious due to relying on expert knowledge.

Although plenty of approaches have been proven to be effective, there are gaps between current hoist scheduling problems and real-world industrial problems. For instance, cyclic hoist scheduling problems aim at computing fixed moving sequences for hoists with the goal of minimizing cyclic times. However, they \cite{che2004single,DBLP:journals/tase/CheLFC14} assume all jobs are available at the loading station waiting for being processed in the beginning, neglecting unexpected arrivals and unfamiliar types of new jobs. Real-time and dynamic hoist scheduling approaches \cite{DBLP:journals/cor/AmraouiE16,yan2017heuristic,zhao2013real} are proposed to plan out even if facing unexpected disruptions. However, they have troubles in  handling large-scale tasks. Therefore, approaches \cite{zhao2013real,DBLP:journals/cor/YanLSM18} made some assumptions to let problems easier to be modeled, such as the locations of the loading and unloading stations, single-track production lines, and a small number of hoists. Nevertheless, those assumptions are not applicable to all real-world industrial scheduling problems. In other words, mathematical models have to be rebuilt and optimized as long as environments change, intensifying contradictions in real-time scheduling problems between states used for calculations and actual states after running time. Those limitations restrict their real-world application. 

Similar to hoist scheduling problems, Automated Planning (AI planning) is adept at computing valid plans with correct logic based on domain knowledge and instance descriptions \cite{DBLP:journals/ai/ZhuoK17,DBLP:journals/ai/JinZXWK22}. Differently, AI planning relies on universal descriptions of domains, including preconditions and effects, focusing on logical relations and rules of updating between actions. Therefore, given initial assignments and achievable goals, planners can compute valid action sequences to guide agents to achieve goals, different from constructing mathematical models based on specified constraints done by previous hoist scheduling approaches. \hankz{It is, however, challenging to complete large-scale tasks of real-world industrial scheduling with off-the-shelf planning techniques. A potential effective way is to integrate planning techniques into mathematical models-based scheduling framework, with planning techniques dealing with high-level logical relations and mathematical models-abased scheduling techniques dealing with low-level detailed constraints, which benefit from each other in solving the industrial scheduling problem effectively and efficiently.}


To do this, we need to deal with three main issues:
\begin{itemize}
    \item The first issue is that products are dynamically added into production lines, incurring uncertain conflicts. 
    \item The second is that environments may unexpectedly change since some tanks may be broken, which means solutions have to be rebuilt with respect to both previously executed actions and current changed environments.
    \item The last one is since we need to rebuild new solutions based on a starting state $s_0$ (i.e., a snapshot of the environment) while the production line is running based on old solutions, we need to estimate the starting state $s_0$ in which the production line starts to execute new solutions (after new solutions are successfully rebuilt).
\end{itemize}
Considering the above-mentioned issues, we propose a planning-based approach, \textbf{H}ierarchical \textbf{I}ndustrial \textbf{T}emporal Planner, a.k.a., {\ours}, to handle real-world real-time hoist scheduling problems. We regard hoist scheduling problems as temporal planning problems, \hankz{with concurrent and durative actions. We design the domain model according to real-world electroplating production lines, where} each action is to operate a hoist, such as moving a hoist from a tank to another one. Specifically, we first generate skeleton schedules by neglecting ``detailed'' constraints. With the skeleton schedules, we compute action sequences to complete tasks with a temporal planner. To generate reliable guidance, we construct a hierarchical framework, which constructs sub-goals for planning and estimates the time of recomputing sub-goals by an inference mechanism. The hierarchical framework allows {\ours} to react quickly when unexpected disruptions happen, and handle large-scale searching space. \ignore{To sum up, constructing hierarchical framework is both of building bridges between sketches to detailed plans and dividing large-scale problems into small ones which are easily conquered. }



\ignore{
In the remainder of the paper, we first introduce related works and a formal definition. After that, we present our approach in detail and evaluate our approach by experiments. Finally, we conclude the paper with future work. 
}

\vspace{-1.5mm}
\section{Related Work}
In this section, we introduce related works from three aspects: (1) cyclic hoist scheduling problems, (2) dynamic hoist scheduling problems, and (3) planning-based scheduling problems. We will describe those aspects in detail below.

\vspace{-1mm} 
 \subsection{Cyclic hoist scheduling problem}
Cyclic hoist scheduling problems aim to find cyclic sequences of hoists that minimize the cycle time or resource usage, assuming that all products wait at loading stations in the beginning. Recently, numbers of approaches were proposed, based on two mainstream methods: mathematical models and searching algorithms. For example, researchers \cite{phillips1976mathematical,shapiro1988hoist,liu2002cyclic,zhou2003single,li2013mixed,zhou2012mixed,DBLP:journals/candie/FengCC18} used mathematical models, such as mixed integer programming (MIP), to solve single hoist scheduling problems with multiple tanks. To make approaches closer to real-world production lines, researchers \cite{liu2005efficient,steneberg2013milp,mao2018mixed,laajili2019collision} focused on multi-capacity tanks and multiple hoists. 
On the other hand, searching algorithms, such as branch-and-bound algorithms \cite{che2004single} and evolutionary algorithms \cite{lim1997genetic,manier2008evolutionary,laajili2019genetic,laajili2021adapted} are widely used in cyclic hoist scheduling problems. However, they can not handle unexpected situations and they are not extensible when facing products with new processing crafts. 

\vspace{-2mm}
\subsection{Dynamic hoist scheduling problem}
Compared with cyclic hoist scheduling problems, dynamic hoist scheduling problems assume that products randomly arrive at loading tanks, requiring approaches to reschedule for unexpected changes. Similarly, the mainstream approaches are mathematical models and searching algorithms. For example, researchers \cite{zhao2013real,feng2015dynamic,yan2018dynamic} built mathematical models to handle dynamic single or multiple hoist scheduling problems. 
On the other hand, heuristic-based approaches \cite{yih1994algorithm,lamothe1995dynamic,yan2017heuristic}, branch-and-bound approaches \cite{lamothe1995dynamic,yan2014two,kim2018real} were proposed to search feasible solutions, aiming at higher productivity and better product quality. Considering real-world applications, researchers are interested in real-time hoist scheduling problems \cite{chauvet2000line,sand2004modeling,reddy2018effective}. 
However, dynamic hoist scheduling problems are trapped by large-scale searching space and complex reality constraints, limiting their real-world applications.

\vspace{-2mm}
\subsection{Planning-based scheduling problems}
Automated Planning (AI planning) aims at synthesizing plans according to domain and problem models to transit initial states to goals. In recent years, there have been tremendous progress in making use of AI planning in real-world industrial applications, e.g., planning-based command understanding \cite{DBLP:conf/aaai/AzariaKM16} and dialogue systems \cite{petrick2016using,cohen2020back}. In hoist scheduling problems, AI planning is one of helpful approaches to compute plans and make decisions \cite{dyner2001planning,DBLP:journals/ors/Kallrath02,abioye2021artificial}. For example, many approaches \cite{manier2000constraint,liu2002cyclic,riera2002improved} model hoist scheduling problems with the syntax of Constraint Logic Programming Scheme (CLP) \cite{jaffar1987constraint}, and then compute solutions abiding by the constraints. \citet{micheli2019temporal} regarded industrial scheduling problems as temporal planning problems and made use of classical planning with temporal constraints to solve hoist scheduling problems by heuristic forward searching. However, similar to mathematical models, large-scale searching space limits the applications of planning-based approaches. 




\vspace{-1mm}
\section{Problem Definition} 
In this paper, we define a real-time dynamic hoist scheduling problem by $\mathcal{M} = \langle \Hoist, \Tank, \Operation, \Product, \recipe, \LocationHoist, \LocationProduct  \rangle$, aiming at computing plans to guide hoists $\Hoist$ running on rails beyond tanks $\Tank$ to transport products $\Product$ between tanks. 

$\Hoist = \{ \hoist_0, \dots, \hoist_m \}$ is a set of hoists, \hankz{each of which} can perform three actions: move, load, and unload a product. Hoists run on rails beyond numbers of tanks. The location of $\hoist_i$ is denoted by $\LocationHoist_i$, which is a symbol, either ``+'' or ``-'', joint with a tank, where ``+'' indicates that the hoist is above a tank and ``-'' indicates that the hoist is at a bottom position. A hoist $\hoist_i$ is restricted to only move within $\TankRegion_i = \langle \tank_x, \dots, \tank_y \rangle$ \hankz{($x<y$)} including a sequence of neighboring tanks. We denote the transportation time for a hoist moving from $\tank_m$ to $\tank_n$ by $\moveduration(m,n)$, and lifting time by $\updownduration$, respectively. 

\hankz{$\Tank = \{\tank_0, \tank_1, \dots, \tank_n\}$ is a set of tanks, each of which belongs to one of the three types: loading, unloading, and processing tanks.} Specifically, loading tanks are used to store products waiting for being processed, processing tanks are used to process products, and unloading tanks are used for storing products which have completed all required procedures. The usage of tank is defined by $\OperationOfTank(\tank_i) = \operation_x$, where $\operation_x \in \Operation$. $\Operation$ is a set of operations $\Operation = \{ \operation_0, \operation_1, \dots, \operation_k \}$, e.g., water washing, loading, and exchanging.  

$\Product = \{ \product_0, \dots, \product_z\}$ is a set of products, which are in loading tanks, waiting for being dispatched to tanks according to their processing recipe. \hankz{They will finally be put into} unloading tanks after all procedures have been done. For a product $\product_i$, we define its location by $\LocationProduct_i$, which can be a tank or a hoist, indicating a product is in a tank or grabbed by a hoist. The recipe is defined by $\recipe_i = \langle \sigma_i, \underline{\duration_i}, \overline{\duration_i}\rangle$. $\sigma_i = \langle \sigma_{i}^0, \sigma_{i}^1,\dots \rangle$ is an operation sequence which requires to be executed in order, where $\sigma_i^j \in \Operation$. $\underline{\duration_i}$ and $\overline{\duration_i}$ are two sequences constraining processing time of $\sigma$, where processing time of $\sigma_i^j$ is required to \hankz{be} more than $\underline{\duration_i}^j$ and less than $\overline{\duration_i}^j$. 

\hankz{Given $\mathcal{M}$ as input,} the goal in this paper is to compute a plan $\Plan$, guiding hoists to transport products between tanks \hankz{such that} products are processed according to their recipes within the required time limit. 
\hankz{We define a plan $\Plan$} by a sequence of triples $\langle a, \gamma, \duration \rangle$, where $a$ is an action executed by hoist, including an action name and zero or more parameters. 
$\gamma \in \mathbf{R}^{0+}$ is a positive real-valued number indicating \hankz{the starting time of action $a$}. $\duration$ is \hankz{the duration of action $a$.}

\begin{figure}[!ht]
    \vspace{-2mm}
    \centering
    \includegraphics[width=0.42\textwidth]{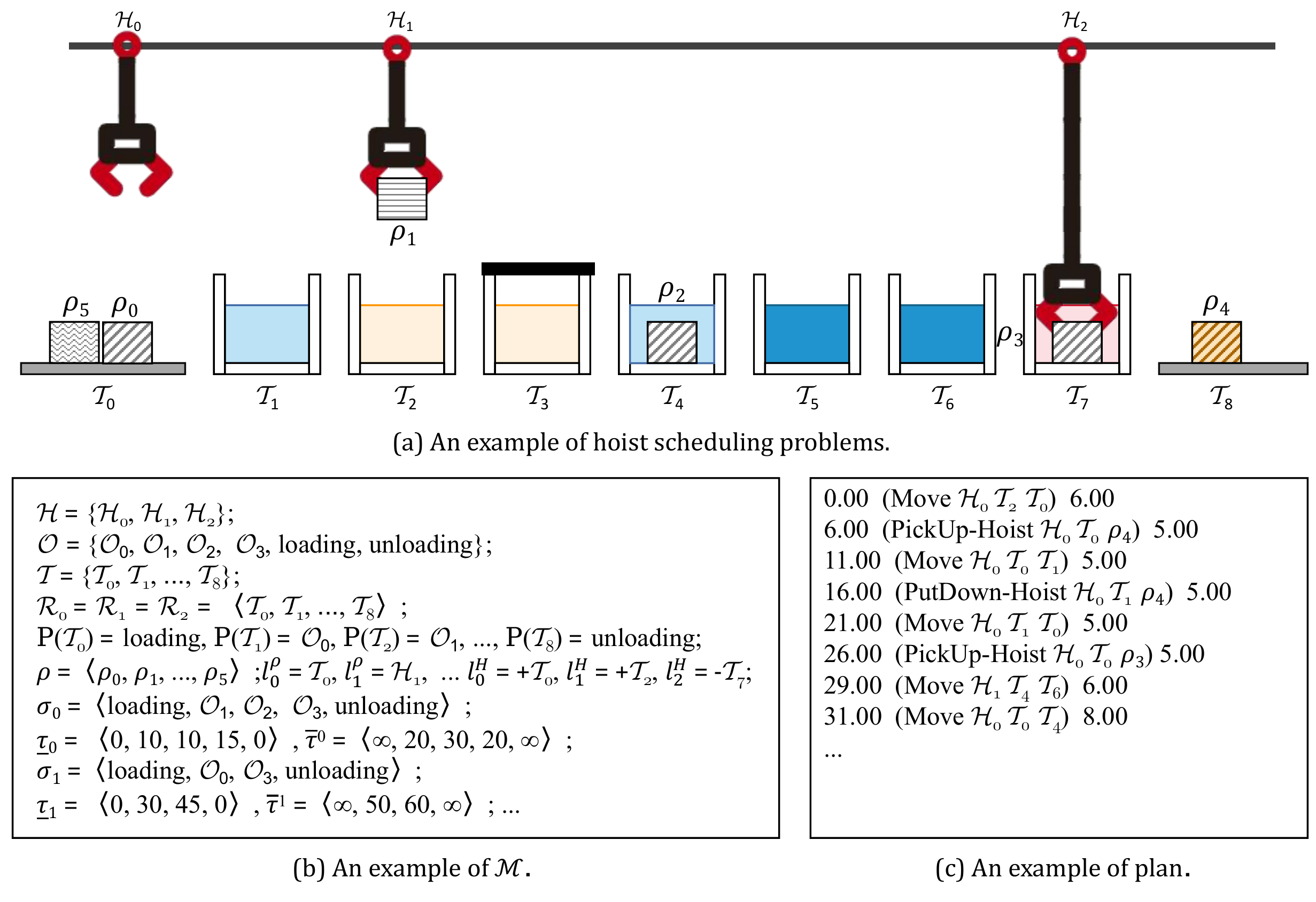}
    \vspace{-3mm}
    \caption{An example hoist scheduling problem. }
    \label{fig:problem_example}
    \vspace{-3mm}
\end{figure}

\emph{  
Figure \ref{fig:problem_example} shows an example in electroplating lines. As shown in Figure \ref{fig:problem_example}(a), the example includes 8 tanks, 3 hoists and 6 products. There are 6 types of operations, including loading and unloading. Tanks can be unavailable, such as $\tank_3$, which can not accommodate any products. Partial description of the example is shown in Figure \ref{fig:problem_example}(b), and the goal is to complete recipes of products following the time requirements. For example, $\hoist_1$ is going to put $\product_1$ into $\tank_5$, whose operation is $\operation_3$, the processing time of the third procedure of $\product_1$ is required to be less than 60 but larger than 45. An example plan in the form of action sequences is shown in Figure \ref{fig:problem_example}(c), where each action is composed of a starting time, an action with zero or more parameters, and a duration. 
}

In this paper, \hankz{according to the real application scenario, we assume the following requirements are satisfied:}
\begin{enumerate}
    \item Tanks can be available or unavailable in the beginning. 
    \item 
    During scheduling, new products may be added to the task. Once they wait in loading tanks, their recipes and arriving time are known.
    \item A product only has one operation at a time.
    \item A processing tank only processes one product at a time, but loading and unloading tanks can contain several ones. 
    \item A hoist only takes one product at a time. 
    \item Tanks keep performing operations. Therefore, the duration of operation is timed as long the product is put into, and it stops when any hoist picks up the products.
    \item The lifting and transportation times are not negligible.
\end{enumerate}

\vspace{-3mm}
\section{Our {\ours} Approach} 

\begin{figure*}[!ht]
    \vspace{-2mm}
    \centering
    \includegraphics[width=0.63\textwidth]{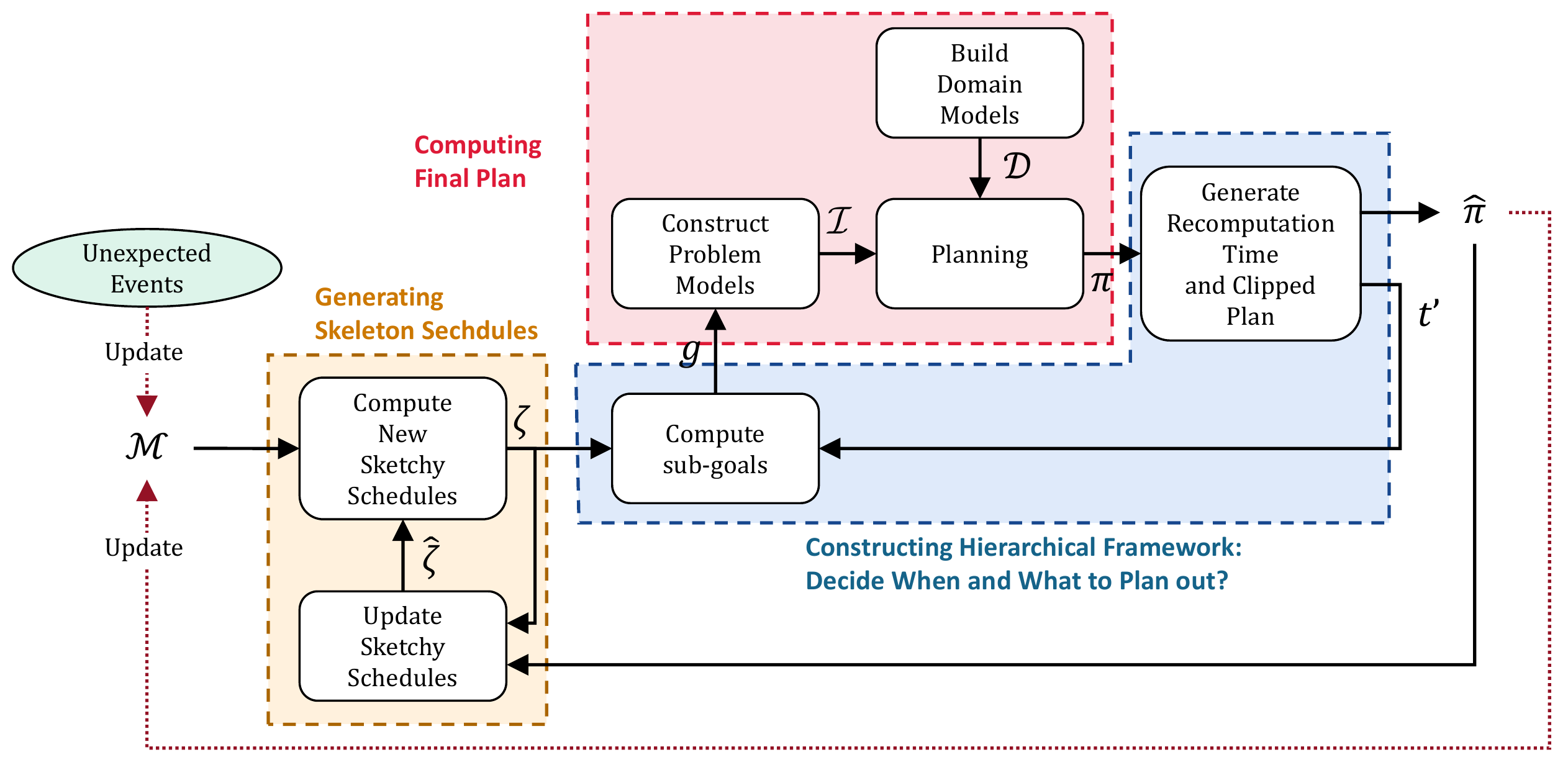}
    \vspace{-3mm}
    \caption{A framework of {\ours}. }
    \label{fig:overview}
    \vspace{-5mm}
\end{figure*}

An overview of {\ours} is shown in Figure \ref{fig:overview}. \hankz{We first} compute skeleton schedules from the whole, ignoring some detailed constraints such as hoist collisions. \hankz{We then} generate sub-goals to divide large-scale scheduling problems into small-scale sub-problems. In this paper, we regard sub-problems as temporal planning problems. \hankz{We manually construct the temporal planning domain model and problem descriptions to represent sub-problems, and exploit a planner, such as TPSHE \cite{TPSHE},} to solve them. To solve large-scale problems, we cut the plan and \hankz{estimate} appropriate times by an inference mechanism to recompute sub-goals. We repeat generating sub-goals and computing sub-plans until all products are \hankz{finished successfully.} 

\vspace{-1.5mm}
\subsection{Generating Skeleton Schedules}
Due to large searching space created by operations sequences, it is hard to solve the whole scheduling problems at once. Therefore, we divide large-scale problems into small-scale sub-problems with sub-goals. However, it is challenging to decide the sub-goals, including computing the order of loading products to be processed, determining the dispatches of hoists and tanks, and minimizing resource usage. To overcome those challenges, we first estimate skeleton schedules in the larger picture ignoring some detailed constraints, which offer the hierarchical framework a rationale to construct sub-goals, instead of forcibly separating the problems and transferring all pressure to a planner. 
Sketch-based search includes two parts: (1) estimating new schedules $\HoistTable$ for new products; (2) updating old schedules $\hat{\HoistTable}$ based on detailed plan $\hat{\plan}$. 

Skeleton schedules are defined by $\HoistTable = \langle \HoistTable_0,\HoistTable_1, \dots \rangle$, recording the occupied time of dispatching hoists and tanks to products transportation and processing. $\HoistTable$ is a set composed of tuples in the form of $\langle \product, \theta, \StartTime,\EndTime \rangle$, where $\product$ is a product and $\theta$ is either a hoist or a tank. A tuple indicates $\product$ is occupied by $\hoist$ or $\tank$ from $\StartTime$ to $\EndTime$. We utilize a sequence $\psi$ to include all starting time $\StartTime$ in $\HoistTable$, from small to large. 

\vspace{-1.5mm}
\subsubsection{New Skeleton Schedules Generation}
When detecting new products not being included in the past estimated schedules $\hat{\HoistTable}$, we first estimate new schedules with the premise of not disturbing $\hat{\HoistTable}$, defined by $\HoistTable = \texttt{Schedule}(\mathcal{M},\hat{\HoistTable})$. 

As for a product $\product$, its recipe and location are defined by $\recipe = \langle \sigma, \underline{\duration}, \overline{\duration}\rangle$ and $\LocationProduct$ respectively. We alternatively compute available hoists to transport the products and tanks to process them, ignoring detailed constraints according to the starting time in $\psi$. Starting from $\psi_0$, we compute all products which can join in the production lines, record them into $\hat{\HoistTable}$, and update $\psi$. When no more products could be dispatched at $\psi_0$, we continue estimating whether the rest products could join at $\psi_1$ and repeat those procedures until all products joining. To compute $\hoist$ and $\tank$ for $\sigma^i$ at $\psi_i$, we search following: 
\vspace{-1mm}
\begin{align}
    &\text{Minimize } (1 + m) \omega
    \label{equation:mathemathical}
\end{align}
\vspace{-2mm}
subject to
\begin{align}
    \omega = \moveduration(\LocationHoist,\LocationProduct)+\updownduration+\moveduration(\LocationProduct,\Tank)
    \label{equation:omega}
\end{align}
where $\omega$ is the least transportation and lifting time, and $m$ is a huge penalty if any following aspects are satisfied: 
\begin{itemize}
    \item $\OperationOfTank(\tank) \neq \sigma^i$.
    \item There is a tuple $\langle \product', \hoist, \StartTime',\EndTime' \rangle \in \HoistTable$ letting $\{y_1| \StartTime' \leq y_1 \leq \EndTime' \} \cap \{y_2|\psi_i \leq y_2 \leq \psi_i+\omega \} \neq \emptyset$. 
    \item There is a tuple $\langle \product', \tank, \StartTime',\EndTime' \rangle \in \HoistTable$ letting $ \{y_1|\psi_i \leq y_1 \leq \psi_i+\omega \} \cap \{y_2| \psi_i+\omega \leq y_2 \leq \psi_i+\omega+\underline{\duration}_i \}
    \neq \emptyset$. 
\end{itemize}
Otherwise, $m=0$. Intuitively, we aim to locate the nearest unallocated hoists and tanks to transport products to be processed with required operations. We denote the occupied durations by $\langle \product,\hoist,\psi_i,\psi_i+\omega \rangle$ and $\langle \product,\Tank,\psi_i+\omega,\psi_i+\omega+\underline{\duration}_i \rangle$. Once we successfully schedule for all procedures, we add the computed occupied durations into $\HoistTable$ and update $\psi$ by inserting new starting time. If we fail computing appropriate tanks and hoists, we continue scheduling for the other products and reschedule for this product next time. 

\subsubsection{Schedules Updating}
Owing to ignoring collisions, predicted skeleton schedules could be different \hankz{from final} plans generated by planners. Therefore, given $\hat{\plan}$, we dynamically update estimated schedules $\HoistTable$ to generate more accurate guidance for later sub-goals generation, defined by $\hat{\HoistTable} = \texttt{Update}(\HoistTable,\hat{\plan})$.
Specifically, we update the starting and ending time of occupied durations in $\HoistTable$ according to $\hat{\plan}$ by executing actions in $\hat{\plan}$ and deferring the occupied durations \hankz{that are} not related to $\hat{\plan}$. Noted that, compared to skeleton schedules, final plans involve unexpected movements of hoists \hankz{based on} constraint satisfactions, which results in differences in the starting time of dispatching hoists to transportation. Therefore, updating schedules focuses on the starting and ending times, instead of recomputing the whole schedules. 

Given $\hat{\plan}$, according to the starting time $\gamma$ and duration $\duration$ of each action, we first update the corresponding $\langle \product, \theta, \StartTime,\EndTime \rangle$ by $\langle \product, \theta,  \gamma,\gamma + \duration \rangle$. Then we update all tuples $\langle \product, \theta', \StartTime',\EndTime' \rangle$ ($\StartTime' > \StartTime$) by $\langle \product, \theta',\StartTime' + \Delta t,\EndTime' + \Delta t\rangle$ where $\Delta t = \gamma + \duration - \EndTime$. After executing $\hat{\plan}$, we utilize $\hat{\HoistTable}$ to include the tuples not updated by actions but deferred. \hankz{Since the updated schedules may involve overlapping occupied durations, we 
use a metric $\nu$ to determine whether any two tuples overlap,} defined by:
\vspace{-2mm}
\begin{align}
    \label{equation:conflicted}
    \nu &= \texttt{Overlap}(\langle \product', \theta', \StartTime',\EndTime' \rangle, \langle \product'', \theta'', \StartTime'',\EndTime'' \rangle) \notag \\
    &= \begin{cases}
    1 &\mbox{if $\EndTime' - \StartTime'' > 0,\product' = \product'' $ or} \\
    &\mbox{ $\EndTime' - \StartTime'' > 0,\theta' = \theta''$, } \\
    0 &\mbox{Otherwise. }\\
    \end{cases}
\end{align}
\vspace{-1mm}
If $\nu$=1, we consider that their durations overlap, where the overlapping duration is denoted by $\delta = \EndTime''-\StartTime'$. We extend the duration of products staying at tanks following the recipes to decrease $\delta$. Specifically, we first filter out $[X_0,\dots,$ $X_m, X_{m+1}, \dots]$ ($X_i = \langle \product'', \theta_i, \StartTime_i,\EndTime_i \rangle$, $\theta_i \in \Tank$, $\StartTime_i \leq \StartTime_{i+1}$, $\StartTime_m = \StartTime''$) from $\hat{\HoistTable}$. If $0 < \EndTime_{m-j} - \StartTime_{m-j} \leq \overline{\duration}_j$, we update all starting time after $X_{m-j+1}$ and all ending time after $X_{m-j}$ by adding $\Delta t$, where $\Delta t = min(\delta,\overline{\duration}_{m-j} - (\EndTime_{m-j}-\StartTime_{m-j}))$. We then update $\delta$ by $\delta = \delta-\Delta t$. Otherwise, we let $j = j+1$. We repeat \hankz{the above-mentioned steps from $j=0$ until no more overlaps happen} or no more durations can be extended. 

\vspace{-1.5mm}
\subsection{Generating Final Plans}
According to sub-goals, we regard them as temporal planning problems, and utilize an off-the-shelf temporal planner, such as TPSHE \cite{TPSHE} exploited in this paper, to compute detailed plans. The reason of using AI planners is that, when environments change, planners allow us to provide new assignments instead of rebuilding the whole models, because logical relations described in domain models are universal. However, it is challenging to directly utilize planners in real-world industrial production lines, because difficulties in large-scale searching space nag AI planners. 
To overcome those challenges, we first build a domain model $\mathcal{D}$ describing common rules in hoist scheduling problems, e.g., a hoist will hold a product after picking it up. In this paper, we regard decomposed sub-problems as temporal planning problems, described by domain and problem models. Problem models $\mathcal{I}$ are built based on sub-goals and $\mathcal{M}$, describing detailed assignments of the environment. Therefore, planner adapts to different environments by updating problem models.
According to domain and problem models, we utilize the planner to compute detailed plans.


Specifically, domain models define universal descriptions made up by action models, where each one is an action taking an amount of time to complete, defined by a tuple of $\langle a, d(a), \pre(a), \eff(a) \rangle$. $a$ is an action name with zero or more parameters. $d(a)$ is either a fixed value or inequalities, indicating the time of executing action. $\pre(a)$ is a set of preconditions, each of which is a logical and temporal expression which must be met in order. Similarly, $\eff(a)$ is a set of effects, describing predicates which will be added into or deleted from current states, or how the variables will be updated. Problem models are composed of initial states and goals, where initial states include grounding predicates and initial assignments of variables. Goals are grounding predicates requiring to be achieved. 

\begin{figure}[t]
    \centering
    \includegraphics[width=0.42\textwidth]{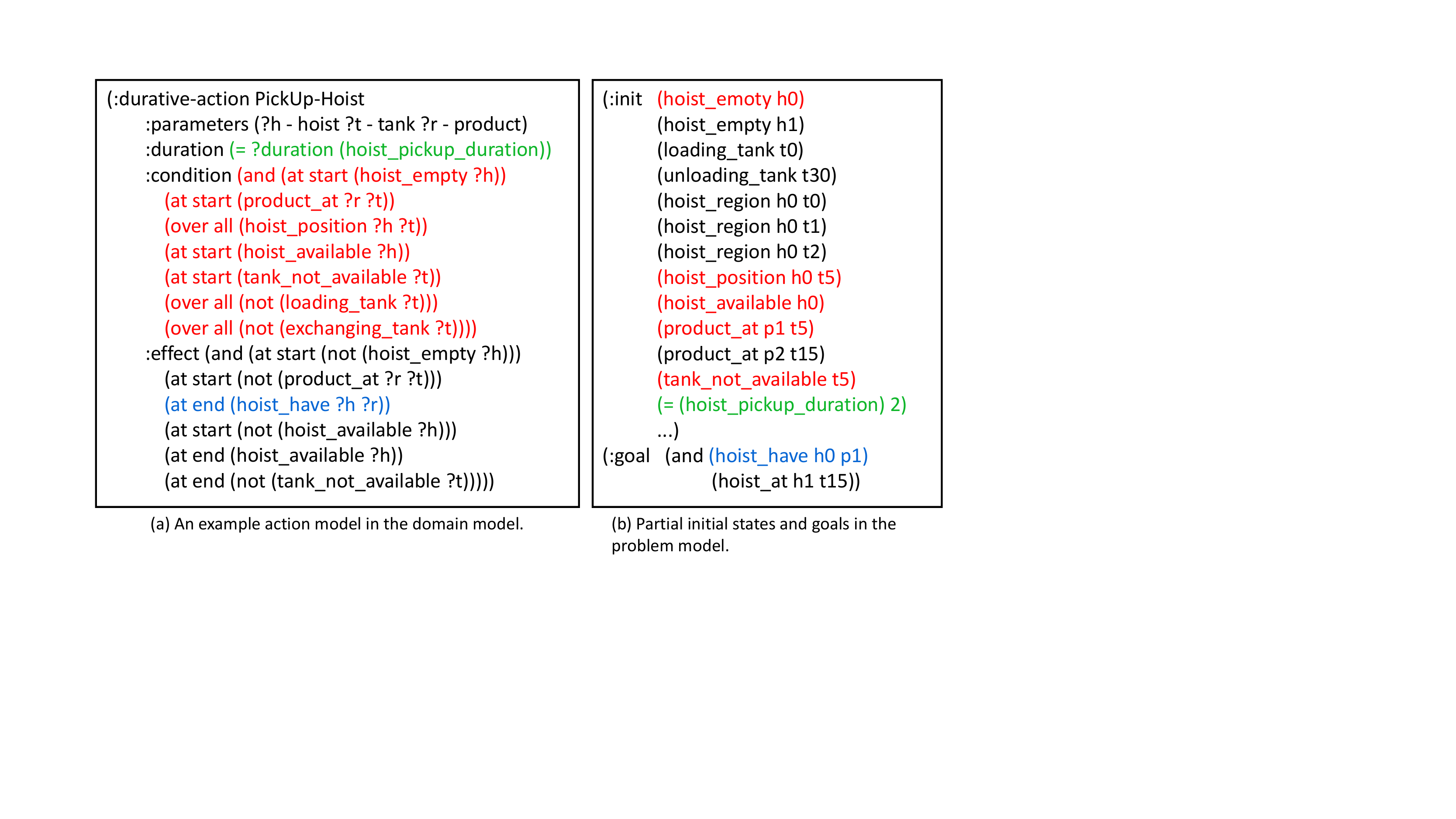}
    \vspace{-4mm}
    \caption{An example of the domain and problem model}
    \label{fig:example_pddl}
    \vspace{-5mm}
\end{figure}

\emph{Figure \ref{fig:example_pddl}(a) shows an example action model\footnote{For space limitations, the whole models are in the supplementary.} ``PickUp-Hoist(?h-hoist ?t-tank ?r-product)''. 
Its duration is defined by``(hoist\_pickup\_duration)''. Seven preconditions are required, such as the hoist must hold nothing at the beginning of executing the action (``(at start (hoist\_empty ?h))''), and the hoist must stay at tank ``?t'' during execution (``(over all (hoist\_position ?h ?t))''). During executing the action, six effects are added in or deleted from the state. For example, ``(product\_at ?r ?t)'' will be deleted from the state once the action is executed, and ``(hoist\_have ?r ?t)'' will be added into the state after executing the action. Figure \ref{fig:example_pddl}(b) shows parts of initial states and goals, where preconditions of action ``PickUp-Hoist(h0 t5 p1)'' are involved, and its duration is 2. After executing it, ``(hoist\_have h0 p1)'' will be achieved. }

According to $\mathcal{D}$ and $\mathcal{I}$, we utilize a temporal planner to plan out a solution $\plan$, the running time is denoted by $\epsilon_r$. To splice the sub-plans, we let all start time in $\plan$ be $\gamma = \gamma + \epsilon$, where $\epsilon$ is the starting time of sub-plans. 

\vspace{-2mm}
\subsection{Constructing Hierarchical Framework}
To build bridges between skeleton schedules and detailed plans, we construct a hierarchical framework, exploring appropriate time to recompute sub-goals to guide refined searching. Therefore, building hierarchical framework includes two parts: (1) generating sub-goals according to skeleton schedules; (2) computing appropriate time to regenerate sub-goals. 

\subsubsection{Sub-Goals Generating}
In real-world scheduling problems, searching solutions for a whole procedure sequence creates a huge searching space, which is hard to model and compute solutions. Therefore, we generate sub-goals to downsize the scale of problems for efficiency and increase quality of solutions. Specifically, given skeleton schedules $\HoistTable$, current processing time of products $\processingtime$ and $\mathcal{M}$, we build sub-goals for schedules which are estimated to be occupied early.

To generate sub-goals, we first select $\langle \product, \theta, \StartTime,\EndTime \rangle \in \hat{\HoistTable}$ with the smallest starting time $\StartTime$. 
Then, we construct sub-goals in the form of grounding predicates following: 
\begin{itemize}
    \item If $\theta$ is a hoist, we regard ``(hoist\_have $\hoist$ $\product$)'' as a sub-goal, indicating that $\hoist$ should pick up $\product$ after executions.
    \item If $\theta$ is a tank, we first seek out another tuple $\langle \product, \hoist, \StartTime',\EndTime' \rangle$, where $\StartTime'$ is the second smallest starting time:
    \begin{itemize}
        \item if $\EndTime' < \StartTime' + \processingtime + \moveduration(\LocationHoist,\tank)$, where $\LocationHoist$ is the location of $\hoist$ and $\processingtime$ is processing time of the product, we regard ``(hoist\_at $\hoist$ $\tank$)'' as a sub-goal, indicating that $\hoist$ should start to move to $\tank$.
        \item Otherwise, we use ``(product\_at $\product$ $\tank$)'' as a sub-goal, indicating that $\product$ should be put into $\tank$ after executions. 
    \end{itemize}
\end{itemize}

\vspace{-1mm}
\subsubsection{Plan Clipping}
After computing sub-plans $\plan$, a nature way is to repeat executing sub-plans and computing new sub-goals until all products are finished successfully. However, due to requirements of different action sequences of each sub-goal, hoists and products which have performed action sequences are asked to wait until all sub-goals are achieved. 
Therefore, we compute appropriate times to construct next sub-problems, including a regenerating time $\epsilon$ as the starting time of next sub-plan and a recomputing time $\hat{\epsilon}$ for real-world applications to recompute sub-goals. Note that $\hat{\epsilon}$ is to avoid errors created by running time. We then cut the plan according to $\epsilon$ and $\processingtime$, deleting actions in $\plan$ executed after $\epsilon$. At last, we update $\mathcal{M}$ and $\processingtime$ according to the clipped plan $\hat{\plan}$.  

Specifically, we first locate the action $\langle a', \gamma', \duration' \rangle$ achieving the first sub-goal. We compute $\epsilon$ following:
\begin{itemize}
    \item If the achieved sub-goal is ``(hoist\_at $\hoist$ $\tank$)'' or ``(product\_at $\product$ $\tank$)'', $\epsilon = \gamma' + \duration'$. 
    \item If the sub-goal is ``(hoist\_have $\hoist$ $\product$)'', $\epsilon = \gamma' - \moveduration(\LocationHoist,\LocationProduct)$, where $\LocationHoist$ and $\LocationProduct$ are locations of $\hoist$ and $\product$, aiming at letting $\hoist$ reach target tanks in advance to avoid products waiting.
    \item If two neighbor actions $\langle a_1, \gamma_1, \duration_1 \rangle$ and $\langle a_2, \gamma_2, \duration_2 \rangle$ exist, where $\gamma_1 < \gamma_2 < \gamma'$, $\gamma_2 \neq \gamma_1 + \duration_1$ and they both move the same hoist $\hoist$, we let $\epsilon = \gamma_1 + \duration_1 + \updownduration$.
    \item If the achieved sub-goal is to put the last product into an unloading tank, $\epsilon = MAX$, where $MAX$ is a hyper-parameter. 
\end{itemize}
Intuitively, we locate: (1) an action achieves the first sub-goal; (2) two actions move hoists but can not be stitched due to spare times existing. We keep actions whose starting time is less than or equal to $\epsilon$, denote the sub-plan by $\hat{\plan}$, and update $\mathcal{M}$ and $\processingtime$ according to $\hat{\plan}$. If a product $\product_i$ is picked up from a tank, we let $\processingtime_i = 0$. If cutting the plan is because of the second reason, we replace the predicate ``(hoist\_start\_moving $\hoist$)'' with ``(hoist\_stop\_moving $\hoist$)'' after updating states to ensure planners to add the time of starting-up to move hoists. Given $\processingtime$, we determine whether $\hat{\plan}$ is valid by:
\vspace{-2mm}
\begin{align}
    \label{equation:validate}
    \texttt{Validate}(\processingtime,\recipe) = 
    \begin{cases}
    1 &\mbox{if~$\forall \processingtime_i \in \processingtime, \underline{\duration}^j_i \leq \processingtime_i \leq \overline{\duration}^j_i$ }\\
    0 &\mbox{Otherwise}
    \end{cases}         
\end{align}
\vspace{-2mm}

Intuitively, if each $\processingtime_i$ satisfies requirements defined in recipes, where $\underline{\duration}^j_i$ and $\overline{\duration}^j_i$ are the lower and upper time requirements of the current operation for $\product_i$, we consider that the plan is valid. Otherwise, the plan is infeasible.

Next, we compute $\hat{\epsilon}$ to apply {\ours} to real-world real-time industrial production lines, in order to avoid inapplicable solutions because of concurrent running of production lines and computations of {\ours}.
To address the challenge, we get benefits from the ability of AI planning to inference future states, and compute a little bit earlier with future states. Specifically, given $\hat{\plan}$ and $\mathcal{M}$, we can compute new $\mathcal{M}$ and $\processingtime$ after executing $\hat{\plan}$. Given the starting time $\gamma_0$ of the current sub-plan $\hat{\plan}$, and the next regenerating time $\epsilon$, we invoke TPSHE a few seconds early to solve the next sub-problem. $\hat{\epsilon}$ is computed by $\hat{\epsilon} = max(\gamma_0,\epsilon-\alpha)$, where $\alpha$ is a hyper-parameter.
After computing new next sub-plan $\hat{\plan}'$, we let all starting time of actions be $\gamma = \gamma + min(0,\epsilon_r-\alpha)$ to avoid errors when running time of planner is more than $\alpha$, where $\epsilon_r$ is exact running time of TPSHE. And we update $\epsilon = \epsilon + min(0,\epsilon_r-\alpha)$. Note that $\hat{\epsilon}$ is used in real-world applications to recompute sub-goals, and $\epsilon$ is the starting time of the next sub-plan. 

\vspace{-2mm}
\subsection{Overview of {\ours}}
\vspace{-3mm}
\begin{algorithm}
\caption{An overview of our Hierarchical Temporal Planning} \label{algorithm:Hit_overview}
\textbf{Input:} $\mathcal{M} = \langle \Hoist, \Tank, \Operation, \Product, \recipe, \LocationHoist, \LocationProduct \rangle$, $\alpha$\\
\textbf{Output:} A plan $\Plan$
    \begin{algorithmic}[1]
    \STATE $\Plan = \langle \rangle, \hat{\HoistTable} = \{ \}, \epsilon = 0, \processingtime = \vec{\textbf{0}}$
    \STATE Build the domain model $\mathcal{D}$
    \WHILE {$\epsilon < MAX$}
    \STATE Compute skeleton schedule $\HoistTable$
    \STATE Generate sub-goals $g$ and model a sub-problem $\mathcal{I}$ according to $\epsilon$, $\processingtime$, $\HoistTable$ and $\mathcal{M}$
    \STATE Utilize a planner to solve ${\mathcal{D}}$ and $\mathcal{I}$, denoted by $\plan$
    \STATE Compute next starting time $\epsilon$ of sub-plans and the clipped plan $\hat{\plan}$, and next recomputing time $\hat{\epsilon}$ according to $\alpha$ and $\plan$
    \STATE According to $\hat\plan$, update $\mathcal{M}$ and $\processingtime$, and compute $\hat{\HoistTable}$
    \IF{$\texttt{Validate}(\processingtime,\recipe)=0 $}
    \RETURN False
    \ENDIF
    \STATE $\Plan = \langle  \Plan | \hat\plan \rangle$
    \ENDWHILE
    \RETURN $\Plan$
    \end{algorithmic}
\end{algorithm}
\vspace{-3mm}

An overview of {\ours} is shown in Algorithm \ref{algorithm:Hit_overview}. First of all, we initialize two empty sets $\Plan$ and $\hat{\HoistTable}$ to indicate plans and skeleton schedules. We initialize the starting time for plans $\epsilon = 0$, and the current processing time of each product $\processingtime = \vec{\textbf{o}}$. (Line 1). Secondly, we build a domain model $\mathcal{D}$ composed of common rules (Line 2). Thirdly, we compute skeleton schedules $\HoistTable$ to estimate occupied duration of products (Line 3). Then we generate sub-goals, model the sub-problem $\mathcal{I}$, and utilize a planner to generate $\plan$ (Lines 5 to 6). Next, given $\plan$, we compute the time to recompute sub-goals $\epsilon$, and use $\epsilon$ to compute $\hat{\plan}$ and $\hat{\epsilon}$ with the help of $\alpha$ (Line 7). According to $\hat{\plan}$, we update $\mathcal{M}$ and $\processingtime$, compute $\hat{\HoistTable}$ (Line 8). If the plan is valid, we insert $\hat{\plan}$ into $\Plan$ (Lines 9 to 12). We repeat computing sub-goals, constructing problem models, and solving problems until $\epsilon = MAX$, indicating all products have been finished successfully (Lines 3 to 13).

\vspace{-2mm}
\subsection{Properties of {\ours}}
{\ours} can be shown to have the following properties:

\noindent \textbf{Theorem 1: (Conditional Soundness)} If the off-the-shelf planner exploited in Algorithm \ref{algorithm:Hit_overview} is sound and $MAX$ is large enough, {\ours} is sound. 

\noindent \textbf{Idea of proof:} 
According to Algorithm \ref{algorithm:Hit_overview}, {\ours} first builds skeleton schedules constrained by Equations (\ref{equation:mathemathical} and \ref{equation:omega}) and computes sub-goals regarding the schedules as heuristics. The schedules dispatch available hoists and tanks without conflicts following the operations defined in recipes, and guarantee that generated sub-goals are valid. {\ours} then utilizes an off-the-shelf planner to complete the details ignored in schedules \hankz{by solving} problems represented by the domain model, initial and goal states, built based on real-world hoist scheduling scenarios and validated by domain experts. Next, according to computed plans, {\ours} computes appropriate time to recompute sub-goals following human-made rules which are validated by domain experts. According to recomputing time, {\ours} cuts the plan, updates states and computes new sub-goals asking {\ours} to push towards next operations. According to Equation (\ref{equation:validate}), the output sub-plans are guaranteed to be feasible to $\mathcal{M}$ satisfying the requirements of processing time. {\ours} repeats constructing sub-goals and solving them until all products are finished successfully, i.e., $\epsilon = MAX$. That is to say, if a problem is solvable and the exploited planner has the ability to compute valid solution for each sub-problem, the output plan is a solution plan for the problem.


\hankz{\noindent \textbf{Theorem 2: (Conditional Completeness)} If the off-the-shelf planner used in Step 6 of Algorithm \ref{algorithm:Hit_overview} is complete and the maximal number of iterations $MAX$ is large enough, {\ours} is complete. 

\noindent \textbf{Idea of proof:} Except Step 6 in Algorithm \ref{algorithm:Hit_overview} that calls an off-the-shelf planner, all other steps can be executed polynomially. If the planner we use is complete, and the maximal number of iterations $MAX$ is large enough to ensure finishing the processing of all products, {\ours} will eventually output a solution to the input problem if there exist solutions to the problems, i.e., the conditional completeness holds.} 

\vspace{-3.5mm}
\section{Experiment}
In this paper, we ran all of our experiments on a machine with 4.400GHz CPU and 16GB RAM. We set the cut-off time to be 180 seconds, and $\alpha$ to be 2. In this section, we evaluate {\ours} with two state-of-the-art methods with different settings:
\begin{itemize}
    \item \citet{feng2015dynamic} constructed MIP models and solved them by CPLEX. We denote the approach by {\shs} and use the first computed feasible solutions. 
    \item We use {\shs} to compute locally optimal solutions within the cut-off time, denoted by {\shsplus}. 
    \item {\tpp} \cite{micheli2019temporal} is a planning-based temporal planner which models industrial problems by temporal planning problems. 
\end{itemize}






In this paper, we evaluate approaches from four aspects: success rate, makespan, CPU time and waiting time. \textbf{Success rate} is the proportion of solved problems. \textbf{Makespan} is the time that elapses from the start of problems to all products completed. \textbf{CPU time} is the time of approaches computing solutions. \textbf{Waiting time} is the time of stopping production lines to wait for computation. We first define five recipes (``loading'' and ``unloading'' are omitted), including operations and required processing time, as shown below: 

\emph{A}: $\langle \operation_1, 25, 55 \rangle$, $\langle \operation_2, 200,550\rangle$, $\langle \operation_3, 80,150\rangle$, $\langle \operation_4,70,$ 

~~~~~$200\rangle$, $\langle \operation_5, 90,160\rangle$, $\langle \operation_6, 200,400\rangle$.

\emph{B}: $\langle \operation_1, 30,60\rangle$, $\langle \operation_2, 60,90\rangle$, $\langle \operation_3, 200,400\rangle$, $\langle \operation_4, 60,$ 

~~~~~$120\rangle$, $\langle \operation_5, 200,400\rangle$, $\langle \operation_6, 30,120\rangle$, $\langle \operation_7, 35,75\rangle$. 

\emph{C}: $\langle \operation_1, 20,50\rangle$, $\langle \operation_2, 400,800\rangle$, $\langle \operation_3, 70,120\rangle$, $\langle \operation_4, 90,$ 

~~~~~$160\rangle$, $\langle \operation_5, 100,120\rangle$, $\langle \operation_6, 70,200\rangle$.

\emph{D}: $\langle \operation_1, 25,55\rangle$, $\langle \operation_2, 90,160\rangle$, $\langle \operation_4, 80,150\rangle$, $\langle \operation_5, 70,$ 

~~~~~$200\rangle$, $\langle \operation_6, 90,160\rangle$, $\langle \operation_7, 150,300\rangle$.

\emph{E}: $\langle \operation_1, 25,55\rangle$, $\langle \operation_2, 200,500\rangle$, $\langle \operation_3, 80,150\rangle$, $\langle \operation_4, 70,$ 

~~~~~$200\rangle$.

Then we use $\mathcal{N}_\hoist, \mathcal{N}_\tank,\mathcal{N}_\product$ to denote the number of hoists, tanks, and products, respectively. Next. we define operations of tanks by $\OperationOfTank(\tank_i) = \operation_i$, except for $\tank_0$ and $\tank_{\mathcal{N}_\hoist-1}$, which are regarded as the loading and unloading tanks. At last, we define lifting time $\updownduration = 5$, and transportation time $\moveduration(\tank_i,\tank_j) = |j-i| + 4$. 
\begin{itemize}
    \item To evaluate the performance of four methods, we first compared them on problems with different numbers of products based on \emph{recipe A}. 
    
    \item To evaluate flexibility when facing unexpected events, we evaluated them on problems with new products randomly joining in. The experiments are built based on \citet{feng2015dynamic}. We generated problems with $\mathcal{N}_\tank \in \{ 8, 10, 12, 14 \}$ and $\mathcal{N}_\hoist = 1$. We first define $n = U(y_1,y_2)$ to randomly selected an integer from $y_1$ to $y_2$. As for each $\mathcal{N}_\tank$, we randomly generated 10 problems, including two sets of products: $\Product$ and $\Product'$. $\Product$ is a set of products waiting at the loading tank in the beginning, where $\mathcal{N}_\product = U(K/2,K)$. $\Product'$ is a set of products joining during running production lines, $\mathcal{N}_{\product'} = U(K/2,K)$. The time to put $\Product'$ at the loading tank is defined by $U(0,T_0)$, where $T_0$ is the makespan of finishing $\Product$ computed by {\shs}. The number of operations is computed by $U(K/2,K)$ and we randomly match operations to tanks, the upper and lower bounds of duration requirements computed by randomly selecting one of the following cases: (1)$ \underline{\duration}_i^j = U(30, 90), \overline{\duration}_i^j = \underline{\duration}_i^j + U(0, 90)$; (2)$\underline{\duration}_i^j = U(90, 150), \overline{\duration}_i^j = \underline{\duration}_i^j + U(0, 60)$; (3) $\underline{\duration}_i^j = U(150, 270), \overline{\duration}_i^j = \underline{\duration}_i^j + U(0, 150)$. 
    
    \item To analysis the limitations, we compared makespan and running time on small-scale problems. We built 8 groups of problems with different $\mathcal{N}_\hoist, \mathcal{N}_\tank$, and $\mathcal{N}_\product$, where each group included 20 random generated problems. Recipes were randomly selected from \emph{recipes B, C, D and E}.
    
    \item At last, to evaluate the performance of {\ours} in real-world industry, we did experiments with two recipes used in real-world electroplating production lines. 
\end{itemize}

\vspace{-3mm}
\subsection{Experimental Results }

\subsubsection{Results of problems with different numbers of products}

\begin{figure}[!t]
\vspace{-4mm}
\setlength{\abovedisplayskip}{0pt}
\setlength{\belowdisplayskip}{0pt}
\centering
\subfigcapskip=-5pt
\subfigcapmargin = .05cm
\subfigure[Makespan]{
    \begin{minipage}[b]{0.21\textwidth}
    \includegraphics[width=\textwidth]{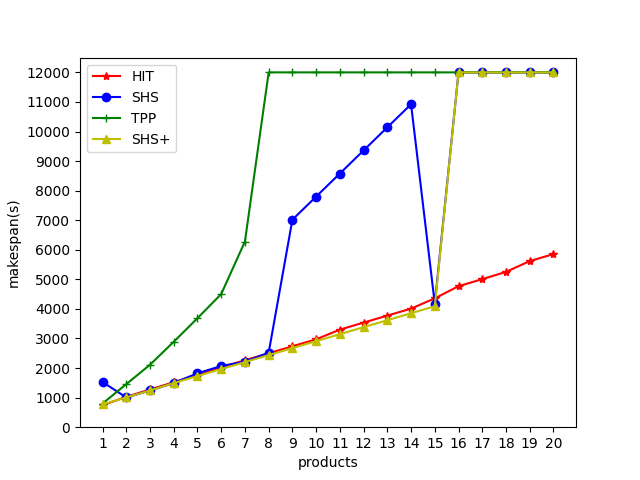}
    \end{minipage}
}
\subfigure[CPU times]{
  \begin{minipage}[b]{0.21\textwidth}
    \includegraphics[width=\textwidth]{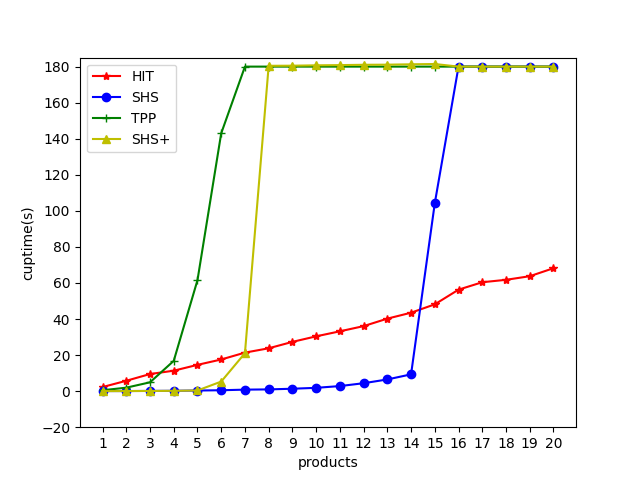}
  \end{minipage}
}
\vspace{-3mm}
\caption{Makespan and CPU time of plans computed on problems with different number of products. }
\label{figure:scalability}
\vspace{-3mm}
\end{figure}

Figure \ref{figure:scalability} shows CPU time and makespan computed by four methods on problems with different numbers of products. As shown in Figure \ref{figure:scalability}(a), compared with {\shs}, {\shsplus} and {\tpp}, {\ours} performs the most steadily, indicating that the makespan is rarely affected by the increasing searching space. On the other hand, although {\shsplus} shows the best performance when $\mathcal{N}_\product < 16$, it can not solve problems with more products. Compared with {\shsplus}, the performance of {\shs} is unstable. {\tpp} shows the worst performance because that forward searching relies on the scale of problems, the growing searching space exponentially increases the difficulties in solving. Figure \ref{figure:scalability}(b) shows the CPU time of four approaches. Similarly, the growth trend of {\ours} is linear, because the hierarchical framework divides large-scale problems into small ones with similar searching space. The running times of the others grow exponentially, indicating that their performances are extremely influenced by the scales. It is noted that, although the makespan of some plans computed by {\shsplus} are the least, {\shsplus} has a large running time in exchange. 

\begin{table}[!t]
\centering
\caption{Results of problems with new products randomly joining in. }
\renewcommand{\arraystretch}{1.1}
\vspace{-2mm}
\label{table:new_products}
\begin{tabular}{llllll}
\toprule
                            & $\mathcal{N}_\tank$ & \makecell[l]{Success\\Rate } & Makespan & \makecell[l]{CPU\\time } & \makecell[l]{Waiting\\time } \\
\midrule
{\shs}     & \multirow{3}{*}{8}           & 1.00         & 2806.60  & 0.60     & 0.43         \\
{\shsplus} &                              & 1.00         & 2427.40  & 1.36     & 1.19         \\
{\ours}    &                              & 1.00         & 2720.60  & 26.37    & 0.00         \\
\hline
{\shs}     & \multirow{3}{*}{10}          & 1.00         & 33661.30 & 6.17     & 2.77         \\
{\shsplus} &                              & 1.00         & 3574.10  & 46.35    & 42.95        \\
{\ours}    &                              & 1.00         & 4396.90  & 47.65    & 0.00         \\
\hline
{\shs}     & \multirow{3}{*}{12}          & 0.60         & 69509.33 & 49.33    & 13.01        \\
{\shsplus} &                              & 0.60         & 4848.83  & 98.46    & 62.22        \\
{\ours}    &                              & 1.00         & 6409.33  & 71.22    & 0.00         \\
\hline
{\shs}     & \multirow{3}{*}{14}          & 0.20         & 59047.50 & 155.45   & 64.42        \\
{\shsplus} &                              & 0.20         & 12749.60 & 180.00   & 91.48        \\
{\ours}    &                              & \textbf{1.00}         & \textbf{7130.00}  & \textbf{78.21}    & \textbf{0.00}        \\
\bottomrule
\end{tabular}\vspace{-5mm}
\end{table}

\begin{table*}[!t]
    \centering
    \footnotesize
    \caption{Results of makespan and running time of plans computed by different methods ($\mathcal{N}_{\product} \leq 10 $).}
    \vspace{-2mm}
    \label{table:results}
    \begin{tabular}{lllllllll}
    \toprule
          \multirow{2}{*}{$(\mathcal{N}_\hoist, \mathcal{N}_\tank,\mathcal{N}_\product)$}
          & \multicolumn{2}{c}{\shs}& \multicolumn{2}{c}{\shsplus}  & \multicolumn{2}{c}{\tpp} & \multicolumn{2}{c}{\ours} \\
        \cmidrule(r){2-3} \cmidrule(r){4-5}  \cmidrule(r){6-7}\cmidrule(r){8-9}
                  & Makespan & CPU time      & Makespan & CPU time     & Makespan & CPU time & Makespan & CPU time      \\ 
        \midrule
(1,6,$<$5)                           & 922.00       & 0.03     & 922.00       & 0.06      & 1365.00      & 1.64     & 932.50       & 6.94     \\
(1,6,$\geq$5)                       & 3341.50      & 8.70     & 2338.00      & 98.22     & 3686.00      & 59.61    & 2419.00      & 23.60    \\
(1,9,$<$5)                           & 1103.25      & 0.14     & 1051.10      & 0.12      & /            & /        & 1202.39      & 10.45    \\
(1,9,$\geq$5)                       & 6219.30      & 1.70     & 2324.50      & 97.18     & /            & /        & 2726.88      & 35.66    \\
(2,6,$<$5)                           & 880.00       & 0.10     & 880.00       & 0.16      & /            & /        & 881.00       & 5.90     \\
(2,6,$\geq$5)                       & 2170.00      & 1.01     & 2170.00      & 1.01      & /            & /        & 2179.00      & 19.16    \\
(2,9,$<$5)                           & 1013.22      & 0.17     & 1013.22      & 0.17      & /            & /        & 1091.10      & 9.63     \\
(2,9,$\geq$5)                       & 4855.60      & 1.85     & 2354.10      & 94.07     & /            & /        & 2502.65      & 31.98    \\
Average                                      & 2563.11      & 1.71     & 1631.62      & 36.38     & /            & /        & 1741.81      & 17.91 \\
    \bottomrule
    \end{tabular}
    \vspace{-4mm}
\end{table*}

\vspace{-1mm}
\subsubsection{Results of problems with unexpected events}

Then we evaluate the flexibility of {\ours}, {\shs}, and {\shsplus} when facing unexpected events. We do not compare with {\tpp} because it can not solve problems with such scales. As shown in Table \ref{table:new_products}, {\ours} successfully compute plans for all problems, where {\shs} and {\shsplus} fail solving some problems when $\mathcal{N}_\tank \geq 12$. Similarly, the CPU time and makespan of plans computed by {\shs} and {\shsplus} grow more rapidly. It is noted that, the makespan of plans computed by {\shs} are very large when $\mathcal{N}_\tank \geq 10$, because {\shs} stops as long as finding feasible solutions, resulting in large starting time of some products. Differently, {\shsplus} takes more time to compute locally optimal solutions, resulting in larger running time. At last, owing to the inference mechanism of {\ours}, computation and execution are simultaneous without termination of production lines, where the other methods have to stop until new solutions. 

\vspace{-1.5mm}
\subsubsection{Analyses}

Next, we compared four methods on 8 groups of problems. As shown in Table \ref{table:results}, the performances of mathematical models ({\shs} and {\shsplus}) depend on the scales of problems. The smaller the scales of problems are, the better performances they have. When $\mathcal{N}_\tank < 5$, mathematical models show a combination of efficiency and high quality. However, {\shs} and {\shsplus} are influenced by enlarging problems scales. {\shs} outputs the first feasible solutions, sacrificing makespan for efficiency when solving larger-scale problems. Differently, it is hard for {\shsplus} to compute the optimal solutions within the cut-off time, which has to be terminated manually. Compared with {\shs} and {\shsplus}, {\ours} are less influenced by changing scales of problems. Averagely, {\ours} is 6\% larger than the makespan computed by {\shsplus}, however the CPU time of running {\ours} is 51\% less than running {\shsplus}. 

\vspace{-1.5mm}
\subsubsection{Case study}
At last, we utilize two real-world electroplating production lines to evaluate {\ours}. The other methods are not experimented because of unacceptable searching space to them. We do not set a cut-off time because of the length of operations. It is noted that {\ours} can simultaneously solve and run production lines without waiting time. Therefore, it can be applied to reality no matter how much CPU time is. 

\begin{figure}[!ht]
\vspace{-3mm}
    \centering
    \includegraphics[width=0.42\textwidth]{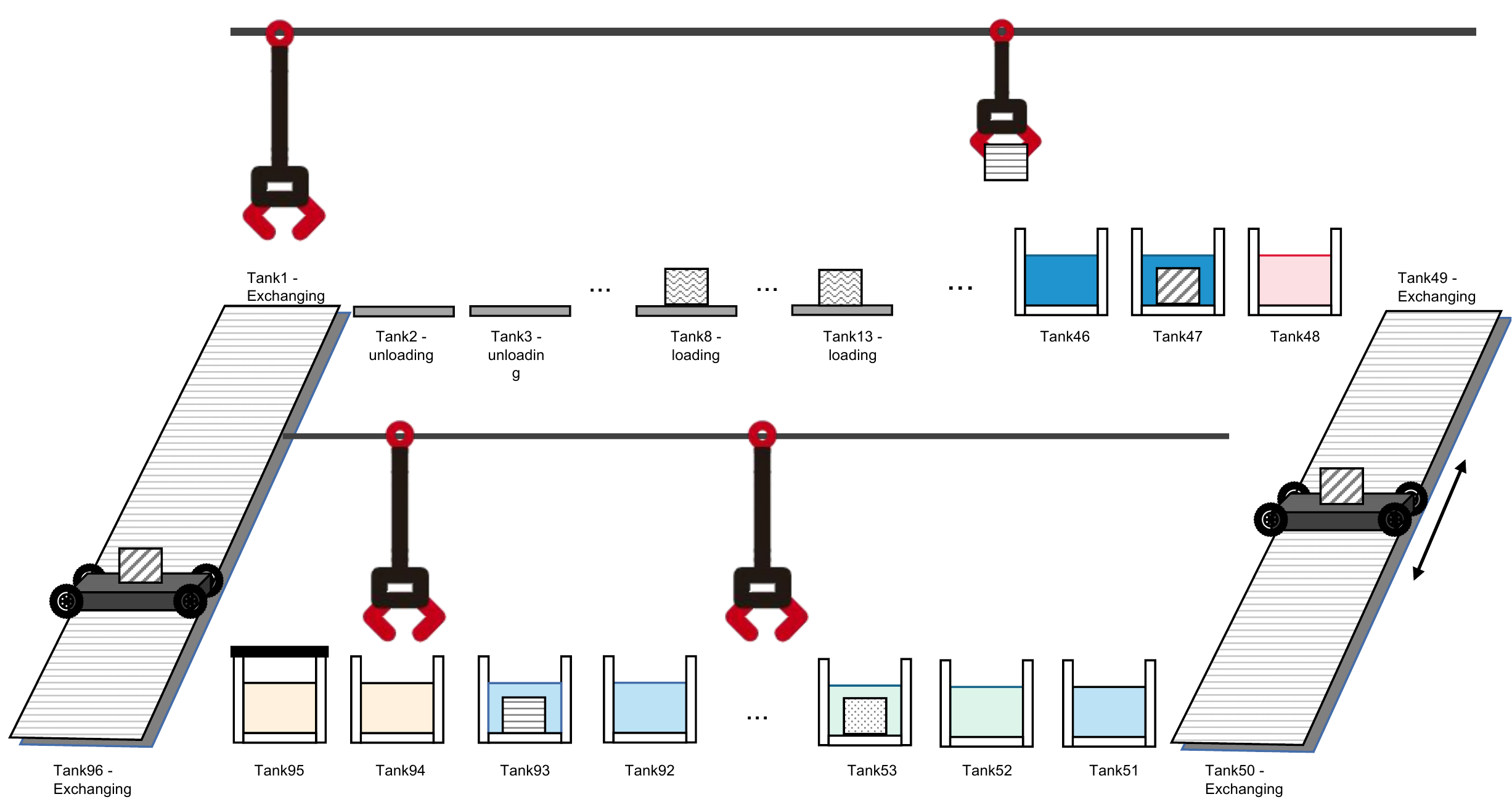}
    \vspace{-1mm}
    \caption{Real-world production lines with 96 tanks. }
    \label{fig:example96}
    \vspace{-3mm}
\end{figure}

The first production line includes 39 tanks, 3 hoists, and 34 operations. The hoists are required to keep safe distance to avoid collisions, where any two hoists should be separated by three tanks. \emph{For example, if $\LocationHoist_1 = +\tank_1$, no hoists can locate at $\tank_2$, $\tank_3$, and $\tank_4$.} The second one includes 96 tanks, 11 hoists, 35 operations, and 2 exchanging carts, as shown in Figure \ref{fig:example96}. Exchanging carts are special hoists which only move forward and back between two tanks connecting two rails. The hoists are required not to locate at the same tank. 

\begin{figure}[!t]
\vspace{-4mm}
\setlength{\abovedisplayskip}{0pt}
\setlength{\belowdisplayskip}{0pt}
\centering
\subfigcapskip=-5pt
\subfigcapmargin = .05cm
\subfigure[Average makespan]{
    \begin{minipage}[b]{0.21\textwidth}
    \includegraphics[width=\textwidth]{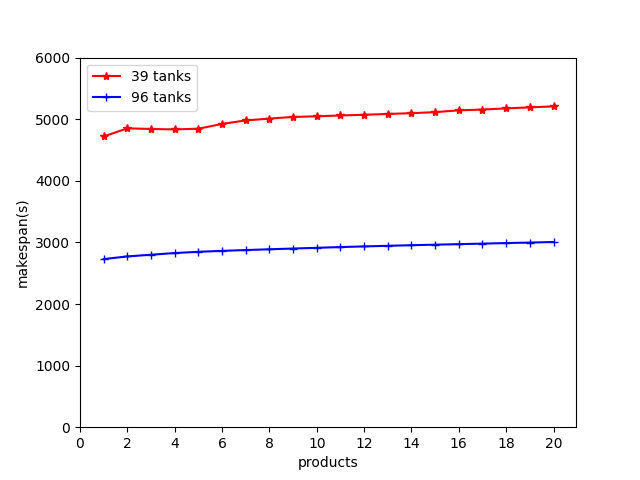}
    \end{minipage}
}
\subfigure[CPU time]{
  \begin{minipage}[b]{0.21\textwidth}
    \includegraphics[width=\textwidth]{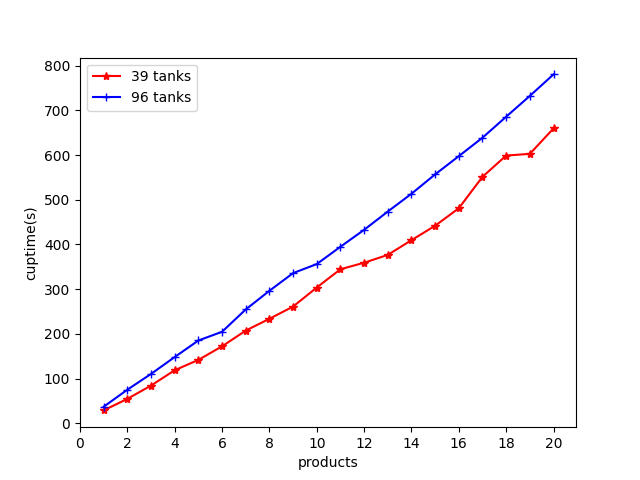}
  \end{minipage}
}
\vspace{-3mm}
\caption{Average makespan and CPU time of plans on problems with 39 and 96 tanks.  }
\vspace{-5mm}
\label{figure:case}
\end{figure}

Figure \ref{figure:case}(a) shows the average makespan of each product, where makespan of a product is defined by the time from loading it until putting it at an unloading tank. As shown Figure \ref{figure:case}(a), the average makespan is stable, indicating that increasing the number of products rarely affects the quality of solutions. Figure \ref{figure:case}(b) shows the running time of solving problems with growing scales, although {\ours} takes much time hierarchically to solve problems, it can simultaneously solve and run production lines without waiting time. Therefore, it can be applied to real-world industry. Partial plans computed for the second production line are shown as below: \\ \\
\begin{tabular}{|c|}
\hline
2667.0~~~~(Move-Hoist hoist11 tank104 tank105)~~~~~~5.00\\
2667.0~~~~(Move-Hoist hoist4 tank43 tank44)~~~~~~~~~~~~5.00\\
2667.0~~~~(Move-Hoist hoist5 tank44 tank45)~~~~~~~~~~~~5.00\\
2672.0~~~~(PutDown-Hoist hoist11 tank105 p0)~~~~~~~~~5.00\\
...\\
\hline
\end{tabular}

\section{Conclusion}
In this paper, we present a novel approach, {\ours}, to solve large-scale real-time dynamic hoist scheduling problems in reality. Specifically, we first generate skeleton schedules as a guide to compute sub-goals, aiming at dividing large-scale problems into small sub-problems. Then we utilize an off-the-shelf temporal planner to compute detailed sub-plans. Next, we compute appropriate times to recompute sub-goals and plan out for real-world industry. Finally, the hierarchical framework allows {\ours} to handle large-scale problems and rapidly respond to unexpected events. By conducting experiments on problems with different settings, the experimental results show the superiority of {\ours}.
Currently, the domain model is built by domain experts. In the future, it would be interesting to investigate automatically learning domain models \cite{DBLP:journals/ai/ZhuoYHL10,DBLP:journals/ai/Zhuo014,DBLP:journals/ai/ZhuoM014,DBLP:conf/aaai/JinMJZCY22} from historical data collected from industry. Besides, it would be also interesting to consider combining skeleton schedules with more mathematical constraints to get less resource usage.


\newpage 
\bibliography{ref}

\end{document}